\documentclass[letterpaper, 10 pt, conference]{ieeeconf}
\IEEEoverridecommandlockouts
\title{\LARGE \bf 
Long-Term Autonomous Ocean Monitoring with Streaming Samples
}
\author{Weizhe Chen, \ Lantao Liu\\
School of  Informatics, Computing, and Engineering\\
Indiana University, Bloomington, IN 47408, USA. \\
E-mail: {\tt\small \{chenweiz, lantao\}@iu.edu}
\vspace{-10pt}
}
\usepackage{graphicx}
\usepackage{lipsum}
\usepackage{bm}
\usepackage{mathtools}
\usepackage{amsfonts}
\usepackage{pgfplots}
\usepackage{cancel}
\usepackage{balance}
\newcommand*\diff{\mathop{}\!\mathrm{d}}
\newlength\figureheight
\newlength\figurewidth
\pgfplotsset{compat=1.14}
\DeclarePairedDelimiter{\ceil}{\lceil}{\rceil}

\begin{document}
\maketitle
\thispagestyle{empty}
\pagestyle{empty}
\begin{abstract}
In the autonomous ocean monitoring task, the sampling robot moves in the environment and accumulates data continuously.
The widely adopted spatial modeling method --- standard Gaussian process (GP) regression --- becomes inadequate in processing the growing sensing data of a large size.
To overcome the computational challenge, this paper presents an environmental modeling framework using a sparse variant of GP called streaming sparse GP (SSGP). The SSGP is able to handle streaming data in an online and incremental manner,  and is therefore suitable for long-term autonomous environmental monitoring.
The SSGP summarizes the collected data using a small set of pseudo data points that best represent the whole dataset, and updates the hyperparameters and pseudo point locations in a streaming fashion, leading to  high-quality approximation of the underlying environmental model with significantly reduced computational cost and memory demand.
\end{abstract}
\section{Introduction and Related Work}
\begin{figure}[t]
    \centering
    \includegraphics[width=1\linewidth]{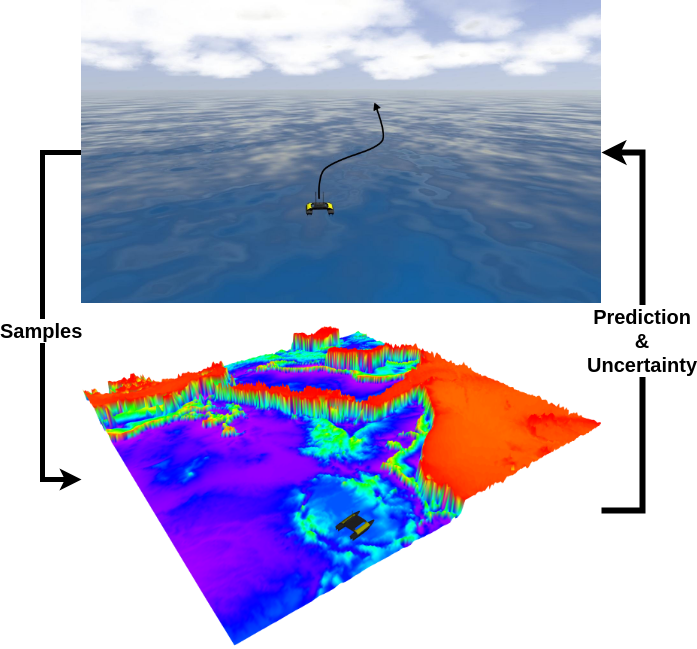}
    \caption{ A typical loop between the sampling robot and the environment. The example here is to estimate the bathymetry of ocean floor. 
    }
    \label{fig:pull_figure}
\end{figure}

To autonomously monitor our aquatic environments such as the oceans, intelligent robotic platforms such as the autonomous underwater vehicles (AUVs) and unmanned surface vessels (USVs) have been increasingly utilized in scientific information gathering missions due to the attractive mobility, flexibility, and adaptivity of these platforms~\cite{dunbabin2012robots}.
Fig.\,\ref{fig:pull_figure} illustrates the interaction loop between a sampling robot and the environment. 
Typically, the robot needs to first compute and plan a sampling path following which environmental samples can be collected. The path is computed based on some learned environmental model, which is usually a probabilistic model of targeted environmental attributes such as the non-uniform salinity of the ocean.
Then, the robot samples the environment and uses the newly obtained samples to update its estimated model, which in turn influences the sampling path computation in the next round.

The sampling path computation for environmental sensing and monitoring is usually formulated as the  \textit{informative path planning}, the objective of which is to maximize the ``informativeness'' of collected samples. 
Such informative path navigates the robot to collect samples from locations of greater importance for better estimating the underlying environmental model.
Oftentimes, the informativeness of a new observation (sampling) is quantified by its entropy, variance reduction of prediction, or mutual information between the sampled and un-sampled spaces.
Representative informative planning methods include, e.g., recursive-greedy algorithms that exploit the submodularity property of mutual information~\cite{hollinger2013active, krause2008near, meliou2007nonmyopic, singh2009efficient}, dynamic programming algorithms that select a set of waypoints with maximum information~\cite{cao2013multi, ma2018data, ma2016information, mishra2018online}, sampling-based motion planning algorithms~\cite{hollinger2014sampling, lan2016rapidly}, Bayesian optimization~\cite{bai2016information, ling2016gaussian, marchant2012bayesian}, evolutionary methods~\cite{popovic2017online},  temporal difference reinforcement learning~\cite{chung2015learning}, and Monte Carlo tree search~\cite{arora2018multi, best2019dec, chen2019pareto, morere2017sequential}.

Most existing informative planners have been built upon probabilistic regression models. 
This also requires the environmental modeling component to be efficient and accurate.
Gaussian process (GP) regression has become the de-facto standard for spatial modeling due to its well-calibrated uncertainty estimate, modeling flexibility, and robustness to overfitting.
Nevertheless, there are still many challenges to be overcome.
For instance, one of the big challenges is the limited scalability to large dataset. 
The time complexity and the storage complexity of a vanilla GP regression with $N$ collected data samples are $\mathcal{O}(N^3)$ and $\mathcal{O}(N^2)$, respectively.
Another challenge is the lack of a principled method to deal with streaming data.
As the robot collects data sequentially, we would like to update prediction and (hyper) parameters of our model in a real-time and incremental fashion.
However, most existing work tackled the problem by initializing the parameters through a pilot survey or prior data, and keeping them fixed during the sampling process \cite{binney2010informative, flaspohler2018near, hitz2014fully, kemna2018board}. 
Unfortunately, fixed parameters will undoubtedly limit the adaptivity of the sampling robot.

This paper presents a sparse variant of GP which is able to handle streaming data in an online and incremental manner,  and is therefore suitable for long-term autonomous environmental monitoring.
Specifically, the predictive distribution, hyperparameters, and the locations of pseudo points (a small set of points summarizing the training data) are updated in real-time, leading to an accurate estimation of the environmental state with reasonable computational cost and memory demand.
This work relates to sparse online GP (SOGP)~\cite{csato2002sparse, ma2018data} and sparse pseudo-inputs GP (SPGP)~\cite{mishra2018online, snelson2006sparse}.
The key difference between this work and our former SOGP-based environmental sampling work~\cite{ma2018data,ma2017informative} is that, instead of employing crafted heuristics for optimizing and selecting a set of representative sparse samples,  we present a principled framework for learning hyperparameters and optimizing the representative pseudo inputs.
Also, the SPGP framework treats the pseudo inputs as additional kernel hyperparameters, which might lead to overfitting when we jointly optimize hyperparameters and pseudo inputs \cite{titsias2009variational}.
Moreover, in SPGP, there is no discrepancy measure between the approximate model and the exact model.
In contrast, the SSGP~\cite{bui2017streaming} models the sparse pseudo points as variational parameters which are optimized by minimizing the Kullback-Leibler (KL) divergence between the approximate GP and the exact posterior GP (see Section \ref{sec:ssgp}).
Such a mechanism prevents the learning algorithm from overfitting and leads to high-quality approximation.

\section{Background}
In this section, we briefly review GP regression and a variational framework for sparse GP regression.

\subsection{Gaussian Process Regression}
Assume we have $N$ training inputs (i.e. sampling locations) and the corresponding real-value outputs (namely, observations) $\{\bm{x}_n, y_n\}^N_{n=1}$. 
In the vanilla GP regression, we assume $y_n = f(\bm{x}_n) + \epsilon_n$, where $f$ is an unknown function and the observation noise is drawn from a Gaussian distribution $\epsilon \sim \mathcal{N}(0, \sigma^2_y)$.
For notational simplicity, $f$ is typically assumed to be drawn from a zero-mean GP prior $f \sim \mathcal{GP}(\bm{0}, k(\cdot, \cdot | \bm{\theta}))$, where $k(\cdot, \cdot | \bm{\theta})$ is the covariance function with a set of hyperparameters $\bm{\theta}$.
After taking the observed data into account, our prior GP can be updated to a posterior GP which is specified by a posterior mean function and a posterior covariance function:
\footnote{We have aggregated the observations into a vector $\bm{y} = \{y_n\}^N_{n=1}$.}
\begin{align*}
    m_{\bm{y}}(\bm{x}) &= K_{\bm{x}N}(K_{NN} + \sigma^2_yI)^{-1}\bm{y}\\
    k_{\bm{y}}(\bm{x}, \bm{x}') &= k(\bm{x}, \bm{x}') - K_{\bm{x}N}(K_{NN} + \sigma^2_yI)^{-1}K_{N\bm{x}'}.
\end{align*}
Here, $K_{\bm{x}n}$ is a $1 \times n$ covariance vector between the test input $\bm{x}$ and the $N$ training inputs, and $K_{N\bm{x}} = K_{\bm{x}N}^\top$.  
$K_{NN}$ is an $N \times N$ covariance matrix on the training inputs, and $I$ is an $N \times N$ identity matrix.
The posterior GP depends on the values of the hyperparameters which can be optimized by maximizing the log marginal likelihood given by
\begin{equation*}
    \log p(\bm{y}) = -\frac{1}{2}\bm{y}^\top K^{-1}\bm{y} - \frac{1}{2}\log |K| - \frac{n}{2} \log(2\pi).
\end{equation*}
Here we use the shorthand $K = K_{NN} + \sigma^2_y I$.

However, the standard GP suffers from poor scalability because it requires matrix inverse operations with time complexity of $\mathcal{O}(N^3)$ and memory (storage space) complexity of $\mathcal{O}(N^2)$.
This computational challenge has led to many sparse approximation paradigms \cite{bui2017unifying, quinonero2005unifying}.
Next, we will introduce a variational framework for the sparse GP.

\subsection{Variational Sparse Gaussian Process}
Variational sparse GP (VSGP) approximates the intractable posterior $p(f|\bm{y}, \bm{\theta})$ through an approximate posterior $q(f)$.
We measure the discrepancy between the approximate posterior and the true posterior using Kullback-Leibler (KL) divergence.
Then we can minimize this quantity so that the approximate posterior is as ``close'' as possible to the exact posterior.
Since $p(f|\bm{y}, \bm{\theta})$ is unknown, computing the KL divergence $\mathbb{KL}[q(f) || p(f|\bm{y}, \bm{\theta})]$ is infeasible.
Fortunately, using Bayes' theorem, we have the following important equation:
\begin{equation}
    \mathbb{KL}[q(f) || p(f|\bm{y}, \bm{\theta})] = \log p(\bm{y}|\bm{\theta}) - \mathbb{ELBO}[q(f), \bm{\theta}],
\end{equation}
where $\mathbb{ELBO}[q(f), \bm{\theta}] = \int q(f) \log \frac{p(\bm{y}, f | \bm{\theta})}{q(f)} \diff f$ is the Evidence Lower BOund (ELBO).
Since $\log p(\bm{y}|\bm{\theta})$ does not depend on $q(f)$, maximizing the ELBO w.r.t. $q(f)$ is equivalent to minimizing the KL divergence, which implies that the approximate posterior gets \textit{closer} to the true posterior.
Furthermore, given the fact that the KL divergence is non-negative, ELBO lower bounds the marginal likelihood $\log p(\bm{y}|\bm{\theta})$ so that it can be used for learning the hyperparameters $\bm{\theta}$.

We assume that the approximate posterior has the form $q(f) = p(f_{\neq \bm{u}}|\bm{u}, \bm{\theta}) q(\bm{u})$, where $\bm{u}$ - the pseudo points - is a small subset of $f$, $q(\bm{u})$ is a variational distribution over $\bm{u}$, and $p(f_{\neq \bm{u}}|\bm{u}, \bm{\theta})$ is the conditional prior of $f_{\neq \bm{u}}$.
This assumption induces cancellation of the uncountbly infinite parts in the equation and provides a computationally tractable lower bound:
\begin{align*}
    \mathbb{ELBO} = \int q(f) \log \frac{p(\bm{y}|f, \bm{\theta}) p(\bm{u}|\bm{\theta})\cancel{p(f_{\neq \bm{u}}|\bm{u}, \bm{\theta})}}{\cancel{p(f_{\neq \bm{u}}|\bm{u}, \bm{u})} q(\bm{u})} \diff f.
\end{align*}
The closed-form expression for the optimal variational distribution can be obtained by maximizing $\mathbb{ELBO}[q(\bm{u}), \bm{\theta}]$ w.r.t. $q(\bm{u})$.
Hyperparameters can also be computed by maximizing $\mathbb{ELBO}[q(\bm{u}), \bm{\theta}]$ w.r.t. $\bm{\theta}$.

This framework assumes that the data arrives in batches.
In such streaming setting, we need to append newly acquired data to a continuously-growing dataset, and then re-train the model with the combined dataset.
In the following section, we shall discuss how to update the approximate posterior and hyperparameters by integrating the information from old approximation and the new data in a mathematically sound way.

\section{Streaming Sparse Gaussian Process}\label{sec:ssgp}
Our goal is to derive the new approximate posterior and marginal likelihood using the old approximation and the new data.
Formally, let $q_{\text{old}}(f)$ be the approximate posterior obtained at the previous step.
According to Bayes' rule, we have
\begin{align}
    q_{\text{old}}(f) &\approx p(f|\bm{y}_{\text{old}}) = \frac{p(f|\bm{\theta}_{\text{old}})p(\bm{y}_{\text{old}}|f)}{Z_1(\bm{\theta}_{\text{old}})}\label{eq:q_old}\\
    p(f|\bm{y}_{\text{old}}, \bm{y}_{\text{new}}) &= \frac{p(f|\bm{\theta}_{\text{new}})p(\bm{y}_{\text{old}}|f)p(\bm{y}_{\text{new}}|f)}{Z_2(\bm{\theta}_{\text{new}})},
    \label{eq:q_new}
\end{align}
where $Z_1(\bm{\theta}_{\text{old}}), Z_2(\bm{\theta}_{\text{new}})$ are the normalizing constants.
Since the new posterior should only rely on the new data $\bm{y}_{\text{new}}$ and the old approximation, we rearrange Eq.\,\eqref{eq:q_old} and substitute $p(\bm{y}_{\text{old}}|f)$ into Eq.\,\eqref{eq:q_new}, which yields
\begin{equation*}
    \hat{p}(f|\bm{y}_{\text{old}}, \bm{y}_{\text{new}}) = \frac{Z_1(\bm{\theta}_{\text{old}})}{Z_2(\bm{\theta}_{\text{new}})} p(f|\bm{\theta}_{\text{new}}) p(\bm{y}_{\text{new}}|f) \frac{q_{\text{old}}(f)}{p(f|\bm{\theta}_{\text{old}})}.
    \label{eq:target_posterior}
\end{equation*}
This new posterior fuses the old approximation $q_{\text{old}}(f)$, the new likelihood $p(\bm{y}_{\text{new}}|f)$, and our priors.
However, we cannot use this as the new approximate posterior, $q_{\text{new}}(f) = \hat{p}(f|\bm{y}_{\text{old}}, \bm{y}_{\text{new}})$, because this recovers exact GP regression and it is intractable \cite{bui2017streaming}.
Therefore, we consider approximation by minimizing the KL divergence between $q_{\text{new}}(f)$ and $p(\bm{y}_{\text{new}}|f)$.

Let $\bm{a} = f(\bm{z}_{\text{old}})$ and $\bm{b} = f(\bm{z}_{\text{new}})$ be the function values at the pseudo-inputs before and after seeing new data.
Note that the number of pseudo points $M_{\bm{a}} = |\bm{a}|$ and $M_{\bm{b}} = |\bm{b}|$ are not necessarily the same, and the new pseudo inputs might be different from the old ones.
This is required when new regions of input space are gradually explored in the environmental monitoring scenario.
The forms of the approximate posteriors are assumed to be $q_{\text{old}}(f) = p(f_{\neq\bm{a}}|\bm{a}, \bm{\theta}_{\text{old}})q_{\text{old}}(\bm{a})$ and $q_{\text{new}}(f) = p(f_{\neq\bm{b}}|\bm{b}, \bm{\theta}_{\text{new}})q_{\text{new}}(\bm{b})$ as that of VSGP.
\renewcommand{\d}[1]{\ensuremath{\operatorname{d}\!{#1}}}
\begin{align*}
    &\underbrace{\text{KL}\Big[q_{\text{new}}(f)\Big|\Big|\hat{p}(f|\bm{y}_{\text{old}}, \bm{y}_{\text{new}})\Big]}_{\text{non-negative}}
    = \underbrace{\log \frac{Z_2(\bm{\theta}_{\text{new}})}{Z_1(\bm{\theta}_{\text{old}})}}_{\text{constant}}\\
    &- \underbrace{\int q_{\text{new}}(f) \left[\log \frac{p(\bm{b}|\bm{\theta}_{\text{new}})q_{\text{old}}(\bm{a})p(\bm{y}_{\text{new}}|f))}{p(\bm{a}|\bm{\theta}_{\text{old}})q_{\text{new}}(\bm{b})}\right] \d{f}}_{\mathbb{ELBO}(q(\bm{b}), \bm{\theta}_{\text{new}})}. \notag
\end{align*}
Since the constant term does not depend on $q(\bm{b})$, maximizing $\mathbb{ELBO}(q(\bm{b}), \bm{\theta}_{\text{new}})$ w.r.t. $q(\bm{b})$ guarantees that $q_{\text{new}}(f)$ gets \textit{closer} to $\hat{p}(f|\bm{y}_{\text{old}}, \bm{y}_{\text{new}})$\footnote{We omitted the subscript of $q_{\text{new}}(\bm{b})$}.
Setting the derivative of $\mathbb{ELBO}(q(\bm{b}), \bm{\theta}_{\text{new}})$ w.r.t. $q(\bm{b})$ equal to $0$ gives us the optimal new approximate distribution $q(\bm{b})^*$.
Also, $\mathbb{ELBO}(q(\bm{b}), \bm{\theta}_{\text{new}})$ lower bounds the online log marginal likelihood\footnote{$Z_2 / Z1 \approx p(\bm{y}_{\text{new}}|\bm{y}_{\text{old}})$} since the KL divergence is non-negative.
This lower bound can be used to learn the hyperparameters and optimize pseudo inputs.
This gives us a principled framework for deploying sparse GP in the streaming setting, providing online hyperparameter learning and pseudo-input optimization. See  \cite{bui2017streaming} for more details.
\section{Experiments}\label{sec:experiment}
Our experimental evaluations aim at answering the following questions about SSGP:
\begin{itemize}
    \item[(\textbf{Q1})] Is SSGP able to achieve competitive accuracy with improved computational complexity and memory usage?
    \item[(\textbf{Q2})] Can it effectively characterize the environment by learning  hyperparameters?
    \item[(\textbf{Q3})] How does the number of pseudo points influence the accuracy and  efficiency?
\end{itemize}

\subsection{Experiment Setup}
\textbf{(Setting)} 
We have conducted extensive evaluations with both synthetic and real-world data, and compared the SSGP with the following baseline methods:
\begin{enumerate}
    \item standard GP regression (GPR) with the whole collected dataset~\cite{rasmussen2003gaussian},
    \item standard GP regression with the most recent $500$\footnote{We keep the last $500$ samples so that the runtime of GPR roughly matches that of SSGP} (GPR500) training data points,
    \item variational sparse GP (VSGP) with the whole dataset~\cite{titsias2009variational},
    \item sparse pseudo-inputs GP (SPGP) with the whole dataset~\cite{mishra2018online, snelson2006sparse}.
\end{enumerate}
The covariance function used throughout the experiments is the squared exponential (SE) kernel with 
learned lengthscales $\ell_d$ (as a hyperparameter):
$$k(\bm{x}, \bm{x}') = \sigma^2_f \exp\left[-\frac{1}{2}\sum_{d=1}^D \left(\frac{x_{d} - x'_{d}}{\ell_d}\right)^2\right],$$
where the number of input dimensions $D$ is $2$ in our case.
The hyperparameters and pseudo inputs are initialized with the same values, and then optimized using L-BFGS-B with the same stopping criteria.
All the methods were implemented on GPflow~\cite{matthews2017gpflow} and run on a standard desktop with a 3.6GHz Intel i7 processor and 16GB of RAM.

\textbf{(Data)} Both the synthetic data and the real-world data are simulated on a $100 \text{m} \times 100 \text{m}$ grid map (discrete scalar field) with the grid resolution of 1m in each  dimension, thus the test set contains $10000$ data points.
The synthetic environment is drawn from a two-dimensional GP with hyperparameters $\{\sigma^2_y:0.01$, $\sigma^2_f:1$, $\ell_1:0.3$, $\ell_2:0.7\}$.
We use the real field Sea Surface Temperature (SST) data provided by National Oceanic and Atmospheric Administration (NOAA).
We adopt a na\"ive lawnmower sampling method to best demonstrate the idea and reduce the impact of the sampling mechanism on the modeling and learning results.
The sampling path allows us to collect $4312$ training data points which are observed sequentially along the path.
Specifically, the sampling robot follows the planned lawnmower path and gathers a small batch of data before updating the GP model.
In this way we can split the entire lawnmower path into $98$ small batches, each of which contains $44$ samples.
All the GP models of different batches will update their predictions, hyperparameters, and the pseudo inputs after receiving each batch of data.
The optimized hyperparameters and pseudo inputs estimated from the previous step will be used as initialization for the next round.

\begin{figure*}[t]
    \centering
    \includegraphics[width=1\linewidth]{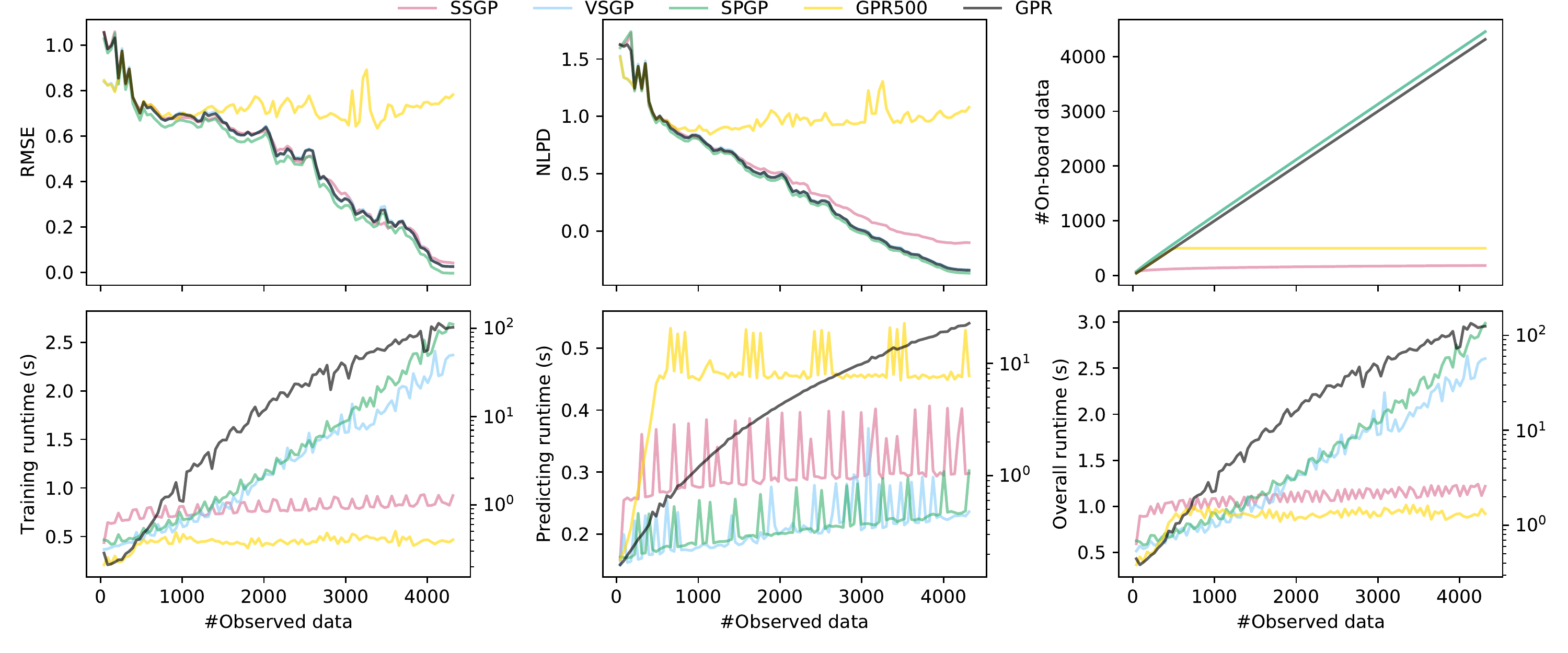}
    \caption{Evaluation results on synthetic data}
    \label{fig:synthetic_metrics}
\end{figure*}
\begin{figure*}[t]
    \centering
    \includegraphics[width=1\linewidth]{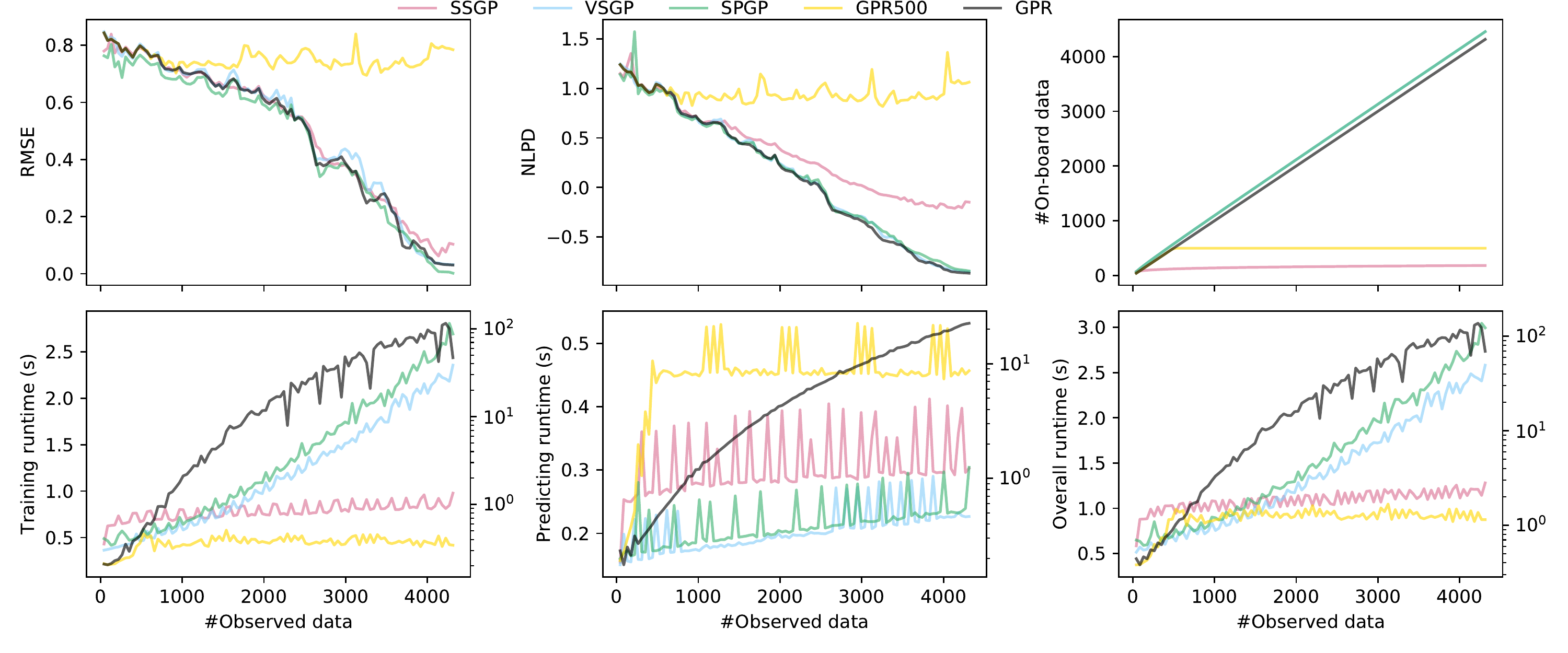}
    \caption{Evaluation results with real-world data}
    \label{fig:real_metrics}
\end{figure*}

\textbf{(Metrics)} To measure the learning performance, we use the Root Mean Squared Error (RMSE) and the average Negative Log Predictive Density (NLPD) on the test data:
\begin{align*}
    \text{RMSE} =& \sqrt{\frac{1}{T}\sum_{t=1}^T (f_t - m_t)^2},\\
    NLPD =& \frac{1}{T}\sum_{t=1}^T -\log p(y_t|\bm{x}_t),
\end{align*}
where $T$ is the number of the test data points, $f_t$ and $y_t$ are the underlying function value and its corresponding noisy observation, respectively; $m_t$ is the posterior mean value, and $p(y_t|\bm{x}_t)$ is the probability density of the predictive distribution.
RMSE only takes a point prediction into account while NLPD penalizes over-confident predictions and under-confident ones.
We also show the numbers of on-board data including training data points and pseudo points to compare the memory usage.
We are revealing that, the training runtime, prediction runtime, and overall runtime demonstrate the computational efficiency of the SSGP method.

\begin{figure*}[t]
    \centering
    \includegraphics[width=\linewidth]{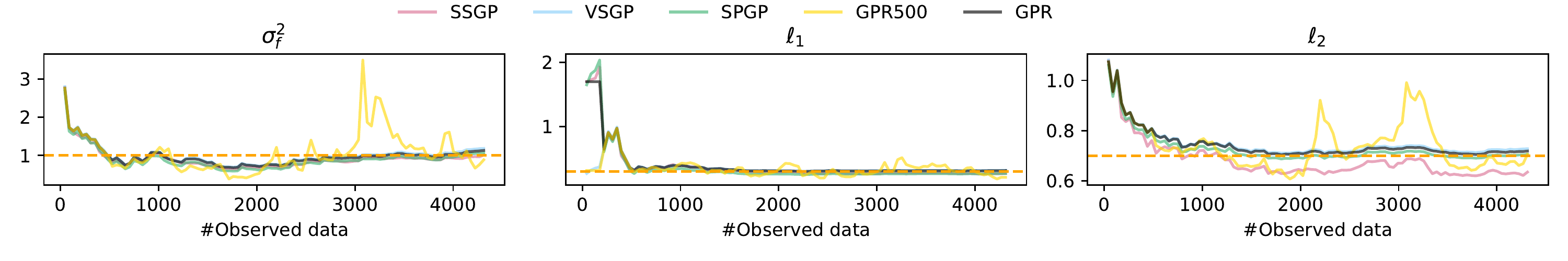}
    \caption{The learned hyperparameters with the ground truth in orange.}
    \label{fig:synthetic_params}
\end{figure*}

\begin{figure*}[t]
    \centering
    \includegraphics[width=\linewidth]{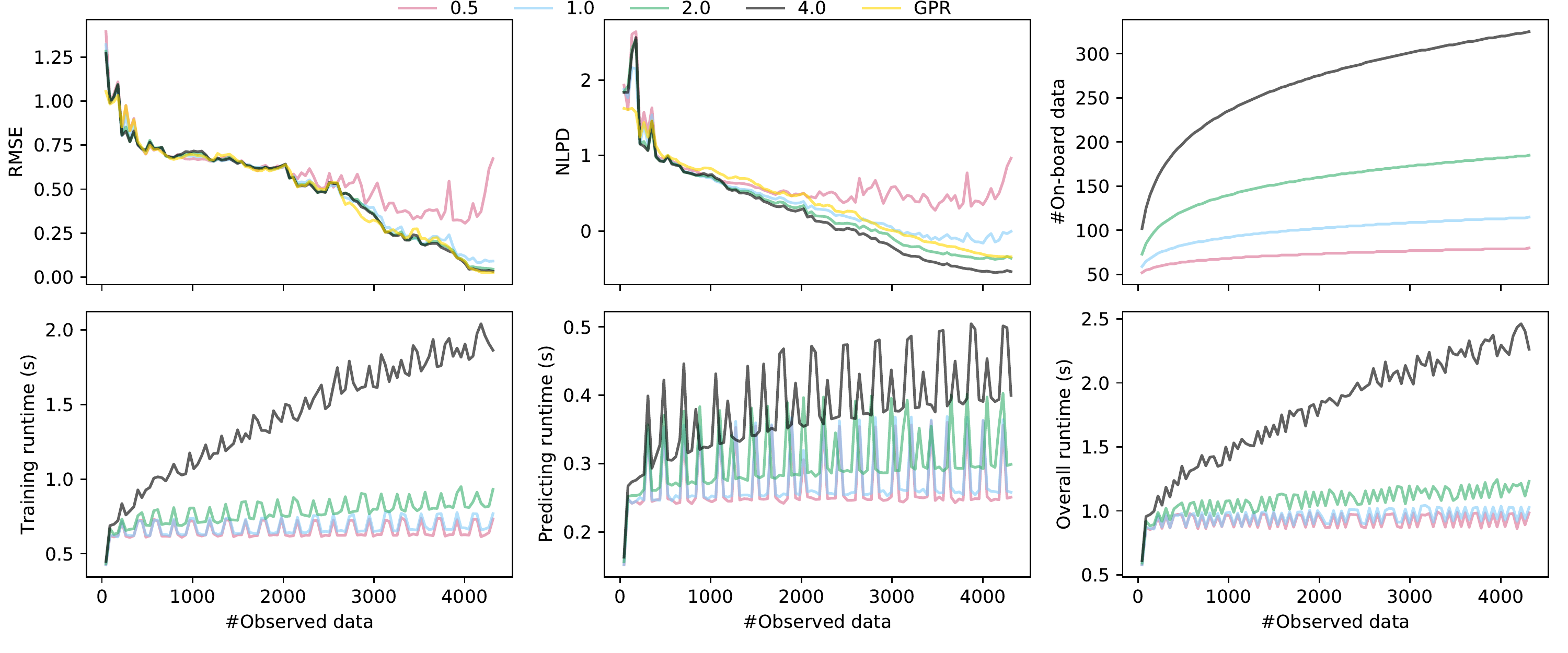}
    \caption{Relationship between the number of pseudo points and the modeling accuracy and efficiency. The values of $\alpha$ are set to be $0.5, 1, 2, 4$ in $M = \ceil{\alpha\log^2N + 1}$.}
    \label{fig:num_pseudo}
\end{figure*}

\subsection{Results}
We first evaluate the algorithms' accuracy, practical runtimes, and memory usage (\textbf{Q1}).
Fig.\,\ref{fig:synthetic_metrics} shows the results of performance evaluation.
Although GPR500 has the fastest runtime, it throws away useful historical information, leading to low accuracy in terms of RMSE and NLPD.
The SSGP, GPR, VSGP, and SPGP have similar RMSEs\footnote{We moved RMSE and NLPD of SPGP down a bit for better visualization.}, but the computational cost of SSGP is much less than the other three.
The overall runtime of SSGP is slightly more than that of GPR500.
(Note, the runtimes of GPR are plotted in log scale.)

The number of on-board data is a reflection of the memory usage.
The SSGP only needs to store the current batch of data and a small set of pseudo points, hence it entails the lowest memory consumption.
This feature is extremely important for long-term autonomous environmental monitoring because 
the robot has limited computation and memory resources while the data size continuously grows.

As noted by \cite{bui2017streaming}, there is a discrepancy between the NLPD of SSGP and that of the other three as more and more data have been collected.
This phenomenon is more obvious on the real-world data as shown in Fig.\,\ref{fig:real_metrics}.
In the next section, we will show that using more pseudo points can mitigate this issue without increasing  much computational cost and memory consumption.

\begin{figure*}
    \centering
    \includegraphics[width=1\linewidth, height=0.83\paperheight]{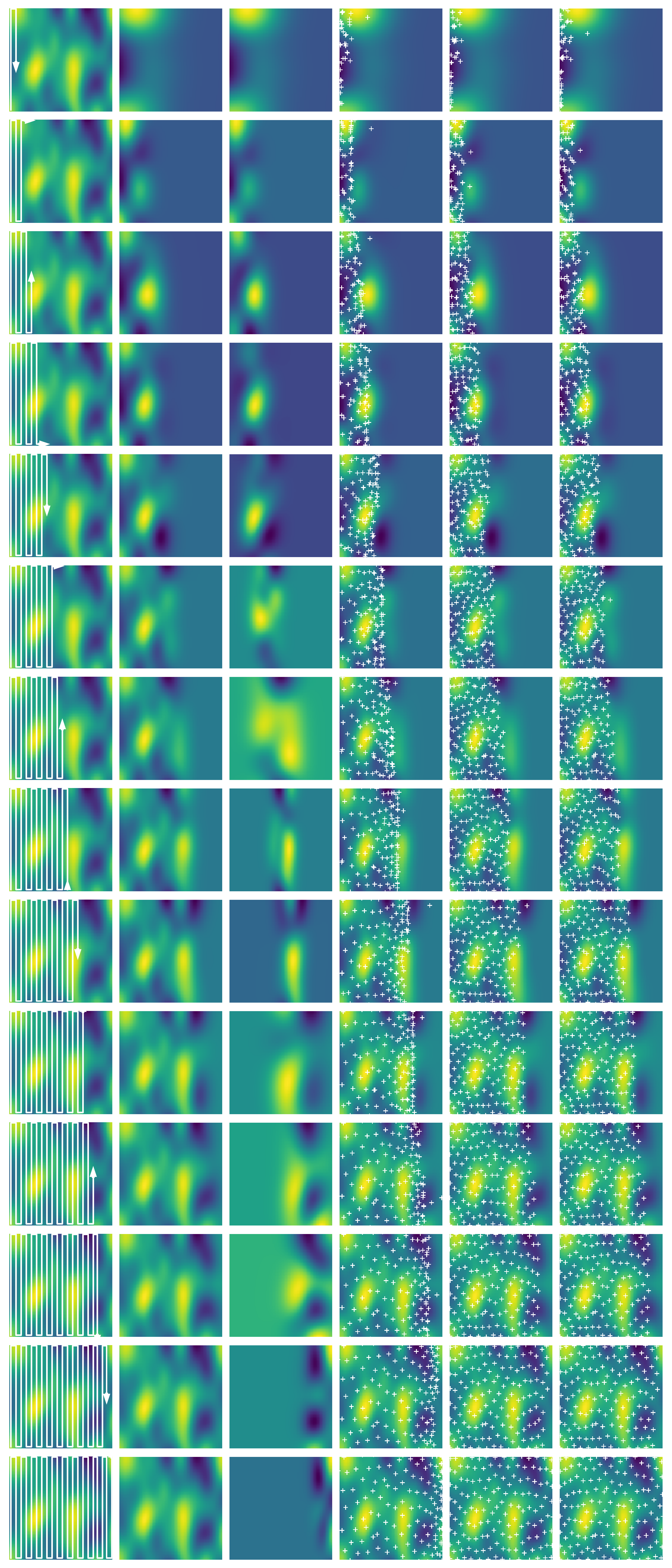}
    \caption{Snapshots of simulation on the synthetic data.
             The first column denotes the environmental ground truth and the robot’s lawnmower sampling path.
             The 2nd to the 6th columns are the predictions of GPR, GPR500, SSGP, VSGP, and SPGP.
             White croses are the pseudo inputs.
             }
    \label{fig:synthetic_snapshots}
\end{figure*}

\begin{figure*}
    \centering
    \includegraphics[width=1\linewidth, height=0.83\paperheight]{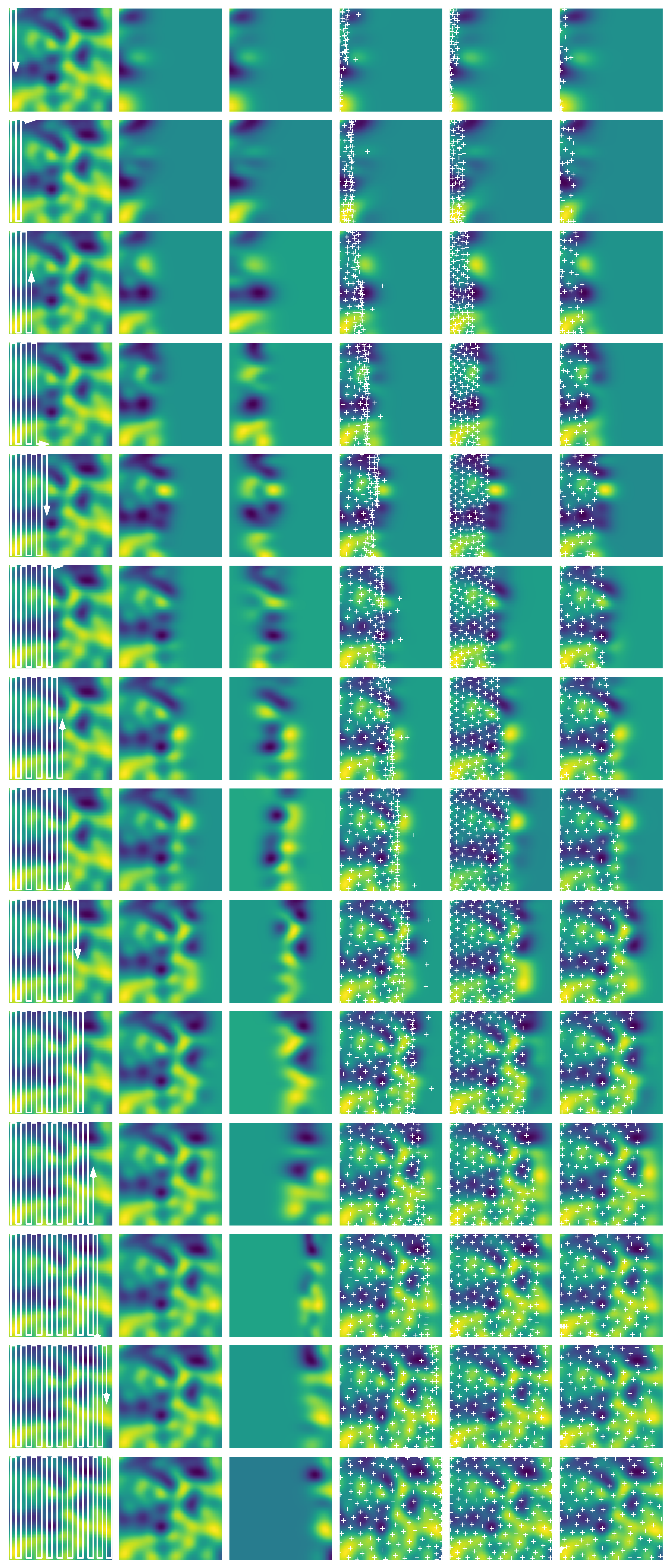}
    \caption{Snapshots of simulation on the Sea  Surface  Temperature  (SST)  data  from National Oceanic and Atmospheric Administration (NOAA). 
             }
    \label{fig:real_snapshots}
\end{figure*}

To investigate the effectiveness of environmental characterization by learning hyperparameters
(\textbf{Q2}), we visualize the learned hyperparameters of all the algorithms together with the ground truth (Fig.\,\ref{fig:synthetic_params}).
The SSGP is able to capture the underlying amplitude ($\sigma^2_f$) and lengthscale along x-axis, though it underestimates the lengthscale along y-axis.
Through the lengthscales of SSGP, we can also infer that the synthetic data varies rapidly along the x direction.
Some snapshots of the modeling process of both datasets are shown in Fig.\,\ref{fig:synthetic_snapshots} and Fig.\,\ref{fig:real_snapshots}.

We then investigate the relationship between the  number  of  pseudo  points  and  the accuracy of the SSGP (\textbf{Q3}).
We summarize the training data using $M$ pseudo points, reducing the computational cost to $\mathcal{O}(NM^2)$.
However, as seen in Fig.\,\ref{fig:synthetic_metrics} and Fig.\,\ref{fig:real_metrics}, the quality of approximation will be limited if the number of pseudo points is not sufficient. (See the increased performance discrepancy between the SSGP and other methods along with the growth of data points where the number of pseudo points keeps unchanged.)
This implies that the number of pseudo points $M$ implicitly depends on the number of training data $N$.
For example, if such dependence is linear, i.e., $M$ scales linearly with $N$, then the computational complexity is still $\mathcal{O}(N^3)$. 
Fortunately, a recent work~\cite{burt2019rates} shows that $M = \mathcal{O}(\log^D N)$ suffices for regression with normally distributed inputs in $D$-dimensions with the SE kernel.
To better demonstrate the trends,  
we ran SSGP with $M = \ceil{\alpha\log^2N + 1}$ pseudo inputs and a set of differing $\alpha$ values to control the number $M$, i.e., $\alpha = {0.5, 1, 2, 4}$. We then plotted the RMSE and NLPD of GPR for comparison.
As shown in Fig.\,\ref{fig:num_pseudo}, when $\alpha$ is $0.5$ or $1$, there are significant gaps between the RMSE and NLPD of SSGP and those of GPR.
When $\alpha=2$, the performance of SSGP matches to that of GPR whilst  the computational times grow with a small margin. When $\alpha=4$, it brings a slight improvement in precision at the cost of more runtime.

\section{Conclusions}
This paper presents an environmental model learning framework using streaming sparse Gaussian process (SSGP).
The SSGP updates the predictive distribution, the hyperparameters, and the pseudo points which summarize the historical data, in a streaming manner.
We have evaluated the performances of SSGP by comparing it against other baseline methods.
Our evaluations show that the SSGP produces competitive prediction accuracy with dramatically reduced  computational cost and memory demand, making it suitable for long-term autonomous environmental monitoring.
We have also empirically investigated how the number of pseudo points influences the learning accuracy and efficiency. Our experimental result reveals that $\ceil{2\log^2N + 1}$ pseudo points are sufficient for regression with $N$ $2$D inputs and SE ARD kernel.
\balance
\bibliographystyle{IEEEtranS}
\bibliography{ref.bib}
\onecolumn

\twocolumn
\end{document}